%% file: main.tex
\documentclass[10pt,twocolumn,letterpaper]{article}

\usepackage{cvpr}              
\usepackage{algorithm}
\usepackage{makecell}
\usepackage{caption} 
\usepackage{algorithmic}
\usepackage{multirow}
\input{preamble}
\definecolor{cvprblue}{rgb}{0.21,0.49,0.74}
\usepackage[pagebackref,breaklinks,colorlinks,allcolors=cvprblue]{hyperref}
\newcommand{\methodname}{{\tt{FEAT}}}
\def\paperID{7418} 
\def\confName{CVPR}
\def\confYear{2026}

\title{From Selection to Scheduling: Federated Geometry-Aware Correction Makes Exemplar Replay Work Better under Continual Dynamic Heterogeneity}

\author{
  Zhuang Qi$^{1}$,
  Ying-Peng Tang$^{2}$,
  Lei Meng$^{1,}$\thanks{Corresponding author}\ ,
  Guoqing Chao$^{3}$,
  Lei Wu$^{1}$,
  Han Yu$^{2}$,
  Xiangxu Meng$^{1}$\\
  $^{1}$School of Software, Shandong University, China \\
  $^{2}$College of Computing and Data Science, Nanyang Technological University, Singapore \\
  $^{3}$School of Computer Science and Technology, Harbin Institute of Technology, China \\
  \texttt{z\_qi@mail.sdu.edu.cn, \{yingpeng.tang, han.yu\}@ntu.edu.sg,} \\ 
  \texttt{\{lmeng, i\_lily, mxx\}@sdu.edu.cn, guoqingchao@hit.edu.cn}
}

\begin{document}
\maketitle
\input{sec/0_abstract}    
\input{sec/1_intro}

\input{sec/2_related}
\input{sec/3_preliminaries}
\input{sec/4_methodology}

\input{sec/5_experiments}

\input{sec/6_conclusion}
{
    \small
    \bibliographystyle{ieeenat_fullname}
    \bibliography{main}
}


\end{document}

%% file: sec/0_abstract.tex
\begin{abstract}
Exemplar replay has become an effective strategy for mitigating catastrophic forgetting in federated continual learning (FCL) by retaining representative samples from past tasks. Existing studies focus on designing sample-importance estimation mechanisms to identify information-rich samples. However, they typically overlook strategies for effectively utilizing the selected exemplars, which limits their performance under continual dynamic heterogeneity across clients and tasks. To address this issue, this paper proposes a \underline{F}ederated g\underline{E}ometry-\underline{A}ware  correc\underline{T}ion method, termed \methodname{}, which alleviates imbalance-induced representation collapse that drags rare-class features toward frequent classes across clients. Specifically, it consists of two key modules: 1) the Geometric Structure Alignment module performs structural knowledge distillation by aligning the pairwise angular similarities between feature representations and their corresponding Equiangular Tight Frame prototypes, which are fixed and shared across clients to serve as a class-discriminative reference structure. This encourages geometric consistency across tasks and helps mitigate representation drift; 2) the Energy-based Geometric Correction module removes task-irrelevant directional components from feature embeddings, which reduces prediction bias toward majority classes. This improves sensitivity to minority classes and enhances the model's robustness under class-imbalanced distributions. 
Experimental results show that \methodname{} outperforms existing methods.

\end{abstract}

%% file: sec/1_intro.tex
\section{Introduction}

\begin{figure}[t]
\centering
\includegraphics[width=1.0\linewidth]{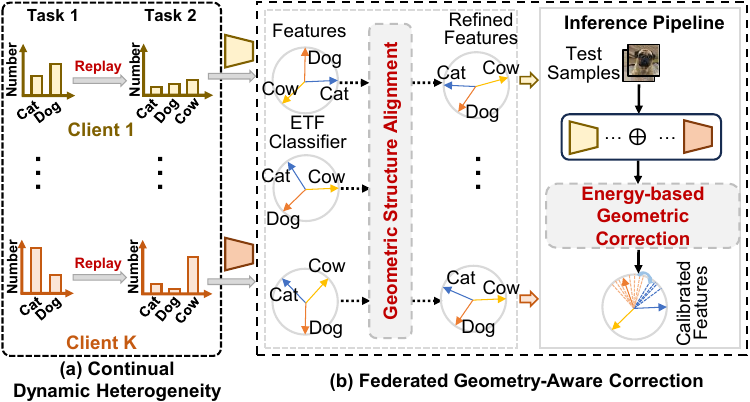}
\caption{
(a) Continual dynamic heterogeneity: clients see different class distributions over time, and each new task is learned with limited replay of past data. (b) Federated geometry-aware calibration: \methodname{} aligns client feature geometry and applies an energy-based correction at inference to calibrate prediction.
} 
\label{fig1}
\end{figure}

Federated learning (FL) is a distributed machine learning framework that enables multiple clients to collaboratively train a shared model without sharing their raw data \cite{hu2024fedmut,qi2023cross,ma2025geometric,yi2023fedgh,hu2024aggregation,liu2024fedbcgd}. It enables periodic model parameter exchanges between clients and the server, where the server aggregates local models from clients to iteratively optimize the global model \cite{fu2025beyond,hu2024fedcross,liuimproving,meng2024improving,yi2024federated,wang2023fedcda}. In real-world FL applications, edge devices are required to handle a continuously evolving sequence of tasks, such as the continual arrival of novel classes, rather than dealing with a static single-task setting \cite{qi2026federated,zhang2025fedagc,yu2024personalized,yu2025handling}. This non-stationary data environment poses a challenge for models to achieve continual learning. Fundamentally, these challenges lie in two aspects: 1) limited local storage hinders the retention of past knowledge; 2) the inherent data heterogeneity across clients, along with the dynamically evolving data distributions, degrades the performance of collaborative learning. These factors lead to severe catastrophic forgetting in federated continual learning (FCL) \cite{yu2024overcoming,shenaj2023asynchronous,li2025unleashing}.

To mitigate catastrophic forgetting, existing studies have widely adopted replay-based methods due to their simplicity and effectiveness. Based on the implementation of replay mechanisms, existing methods can be broadly categorized into two types. (1) Generative-based replay methods utilize generative models, such as Generative Adversarial Networks \cite{navidan2021generative,qi2025federated,chakraborty2024ten}, to synthesize pseudo-samples that resemble historical data \cite{wang2024data,qi2023better,xuankun2025can}. These methods train a generator to approximate the distribution of previously encountered data, which can enhance privacy and reduce memory usage \cite{tran2024text,zhang2023target}. While conceptually appealing, they typically face significant challenges, including the high computational cost associated with training generative models and the potential suboptimal quality of generated samples \cite{wang2024data,tran2024text}. (2) In contrast, exemplar-based replay methods offer inherent advantages in terms of knowledge fidelity, as they avoid the challenges associated with training generative models. They typically store a small portion of the data from previous tasks in local memory and replay it during the training of new tasks \cite{li2024towards,dong2022federated}. Despite the performance advantages demonstrated in existing studies, most of them primarily focus on how to select representative samples, while paying less attention to how to efficiently utilize these limited exemplars under resource constraints \cite{li2025re}.

To address these issues, this paper proposes a federated geometry-aware  correction method for class-incremental learning, termed \methodname{}, which mitigates the optimization bias arising from inter-client data heterogeneity and from the imbalance between majority and minority classes. Figure \ref{fig1} illustrates the main idea of \methodname{}. Specifically, it is designed with two essential components: (1) the geometric structure alignment module enforces angular consistency between local feature representations and globally shared prototypes by distilling relational geometry in the embedding space. By aligning intra-batch feature correlations with those of fixed Equiangular Tight Frame prototypes, it regularizes the local learning dynamics and fosters structurally consistent representations across clients, which can enhance generalization under heterogeneous task distributions. (2) Moreover, to alleviate task-level data imbalance, the energy-based geometric correction module removes task-irrelevant components from the feature space during inference, which reduces overconfidence in majority classes and enhancing the model’s sensitivity to under-represented classes. By jointly leveraging geometric distillation and debiasing, \methodname{} effectively harmonizes local and global objectives, yielding improved performance under heterogeneous and imbalanced cases.

Extensive experiments were conducted on three datasets with varying levels of heterogeneity, including performance comparison, ablation study, sensitivity analysis of hyper-parameters, and case study on the working mechanism of key modules. The results validate that \methodname{} effectively enhances the consistency of representation learning across clients and improves model robustness under long-tailed data distributions. Across all benchmarks, \methodname{} outperforms seven state-of-the-art methods, showing consistent gains in Top-1 accuracy. In summary, this paper makes the following key contributions:
\begin{itemize}[leftmargin=7pt]
    \item This paper reveals two key challenges that persist in exemplar replay-based FCL: the use of replayed data exacerbates inter-client heterogeneity and leads to distributional imbalance between past and current task data.  
    
    \item This study proposes a method orthogonal to exemplar-replay policies, enabling seamless composition with different strategies without changing selection criteria or memory allocation.

    \item The results validate that the prototypical angular distillation improves inter-client feature consistency, the normalized de-biasing classifier alleviates class imbalance during inference, jointly addressing these challenges.


\end{itemize}

%% file: sec/2_related.tex
\section{Related Work}

\subsection{Generative Replay for FCL}
Generative replay trains models such as Variational Autoencoders (VAEs \cite{pinheiro2021variational,girin2020dynamical}) or Generative Adversarial Networks (GANs \cite{goodfellow2014generative}) to approximate the sample distribution of previous tasks, so the learner can revisit pseudo-samples while training new tasks and thus mitigate catastrophic forgetting \cite{babakniya2023don,wang2024data,yang2024federated,nori2025federated,tran2024text,zhang2023target,yu2024overcoming,liang2024diffusion,nguyen2024overcoming}. In FCL, LANDER uses label text embeddings from a pretrained language model to guide data generation \cite{tran2024text}, and GenFCIL uses a lightweight server side generator to synthesize feature representations of old classes from shared class information, reducing privacy risks and memory costs by synthesizing rather than storing data \cite{chen2024general}. However, training reliable generators often requires substantial computation and diverse data, which may be infeasible for edge devices, and generated samples may be of limited quality, which can weaken their ability to prevent forgetting \cite{tran2024text,zhang2023target,tcsvt1}.

\begin{figure*}[t]
\centering
\includegraphics[width=1.0\linewidth]{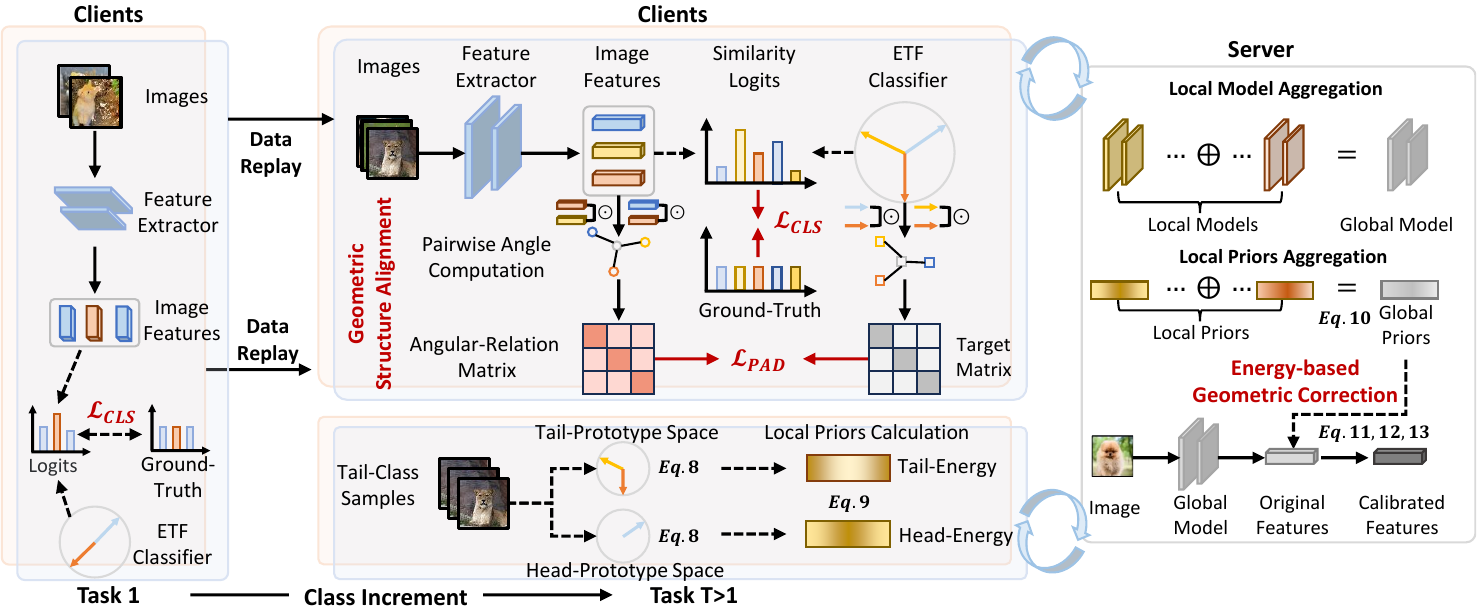}
\caption{Illustration of the proposed \methodname{} framework. It contains two main modules: 1) the Geometry  Structure Alignment module aligns local representations with global ETF prototypes to alleviate inter-client heterogeneity; and 2) the Energy-based Geometric Correction module removes biased components from the feature space during inference. 
} 
\label{fig2}
\end{figure*}

\subsection{Exemplar Replay for FCL}
Given the limits of generative models, many studies use exemplar replay, which stores a small set of representative samples to mitigate forgetting. By keeping real data, exemplar replay avoids the compute cost and sample quality issues of generators, offering a simple and faithful way to retain past knowledge \cite{li2024towards,dong2022federated,liu2025fedadamw,qi2025class,sun2024exemplar,zhang2025few}. For example, Re-Fed \cite{li2024towards} and Re-Fed+ \cite{li2025re} learn a personalized importance model to score local samples and keep those most relevant for reducing forgetting. FedCBDR reconstructs global features of previous tasks in a privacy-preserving way, then guides class-aware and importance-sensitive sampling for balanced replay \cite{qi2025class}. Overall, exemplar replay is practical under limited memory \cite{dong2022federated}, but most works emphasize how to select samples while overlooking how to use the chosen exemplars effectively. As a result, challenges such as client heterogeneity and task-level imbalance remain \cite{li2024towards}, which weakens the performance of the final model \cite{dong2023federated_FISS,li2024cross,dong2026learning,qi2025class}. Other approaches also offer insights for mitigating cross-source bias in FL \cite{koumll,kou2025nips,li2026dvla,koubldl,feng2025neighbor,fengprism}.


\subsection{Equiangular Tight Frame in FL}
Equiangular Tight Frames (ETF) emerge from Neural Collapse theory, where class means form a simplex ETF and classifier weights align with these means, yielding symmetric margins and stable optimization \cite{lu2022neural,papyan2020prevalence}. In FL, several works fix a global simplex-ETF classifier to lessen classifier bias and partially align client representations under non-IID data \cite{li2023no,wu2025enhancing,fan2023federated}. Despite these advances, ETF-based approaches still face feature bias under continual dynamic heterogeneity, as shown in Figures \ref{fig4} and \ref{fig5}. 

%% file: sec/3_preliminaries.tex
\section{Preliminaries}
 
In federated continual learning (FCL), a central server collaborates with $K$ distributed clients to progressively train a global model on a sequence of disjoint classification tasks. Each client $k$ has its own task stream $\{\mathcal{D}_k^{(1)}, \mathcal{D}_k^{(2)}, \dots, \mathcal{D}_k^{(t)}\}$, where new tasks continually introduce unseen classes. To prevent forgetting, replay-based strategies allocate a fixed-size memory buffer of capacity $M$ on each client, which stores up to $N$ representative samples per previous task. When a new task $t$ arrives, the client constructs a memory set $\mathcal{B}_k^{(t-1)}$ by selecting $N$ samples from each of the earlier tasks, ensuring that the total number of stored samples does not exceed $M$. The local training set at round $t$ is then formed by merging the current data with the buffered samples, i.e., $\operatorname{train}^{(t)} = \mathcal{D}_k^{(t)} \cup \mathcal{B}_k^{(t-1)}$. Based on these locally constructed datasets, the global model $\theta_t$ is optimized by minimizing the aggregated training loss across all clients: $\min_{\theta} \sum_{k=1}^K \sum_{(x,y) \in \operatorname{train}^{(t)}} \mathcal{L}(f_k(x; \theta), y)$.

%% file: sec/4_methodology.tex
\section{Methodology}
This section presents a federated geometry-aware calibration method for data replay-based FCL, which performs debiasing across both client-level data heterogeneity and task-level data imbalance, enabling more efficient utilization of limited replay samples. The framework of the proposed method \methodname{} is presented in Figure~\ref{fig2} and Algorithm \ref{alg1}.

\subsection{ Geometric Structure Alignment (GSA)}
\begin{figure}[t]
\centering
\includegraphics[width=1.0\linewidth]{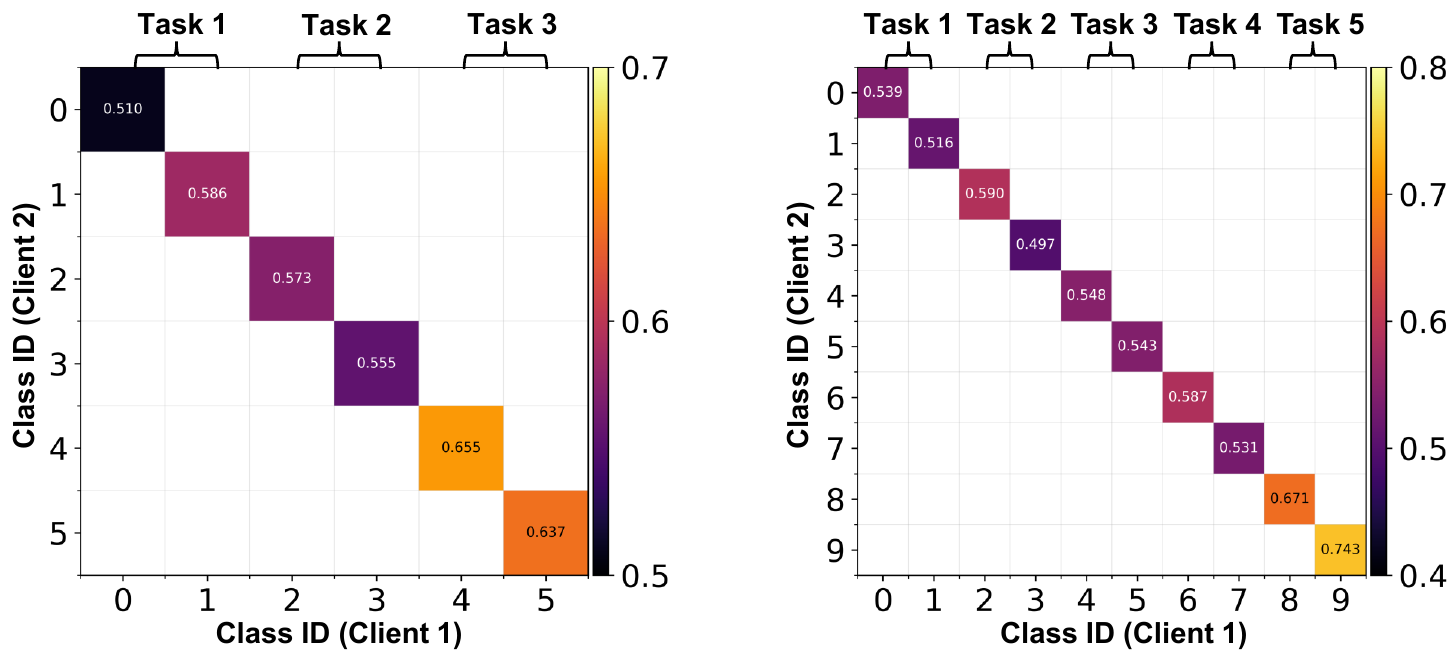}
\caption{Cross-client feature alignment in an incremental 5-task scenario. At Task 3 (left) and Task 5 (right), cross-client alignment for earlier classes is clearly weaker than for current classes.
} 
\label{fig3}
\end{figure}

Despite replay helps retain knowledge from previous tasks, it also introduces continually evolving heterogeneity, making it difficult for clients to maintain a consistent representation space. To alleviate this issue, we leverage a unified geometric prior by adopting the Equiangular Tight Frame (ETF) classifier, which encourages globally consistent class directions across clients \cite{seo2024learning,zheng2025decoupled}. However, under imbalanced class distributions, cross-client alignment of tail classes (previous tasks) remains clearly weaker than that of head classes (current task), as shown in Figure \ref{fig3}. Therefore, the GSA module aims to enforce alignment between the angular structure of the learned features and that induced by their ETF prototypes, enabling a more robust and class-balanced representation space.
Specifically, at incremental task $t$, we adopt a ETF over the currently observed classes $\mathcal{C}_t$ with size
$C_t=|\mathcal{C}_t|$. Let
$W_t=[\mathbf{w}_c]_{c\in\mathcal{C}_t}\in\mathbb{R}^{d\times C_t}$ denote the
lETF prototypes for these $C_t$ classes (with $d\ge C_t$),
\begin{equation}
\small
W_t
=\sqrt{\frac{C_t}{C_t-1}}\;U_t\!
\left(I_{C_t}-\frac{1}{C_t}\mathbf{1}\mathbf{1}^{\top}\right),
\label{eq:etf_construct_t}
\end{equation}

\noindent where $U_t\in\mathbb{R}^{d\times C_t}$ satisfies $U_t^\top U_t=I_{C_t}$,
$I_{C_t}$ is the identity, and $\mathbf{1}$ is the all-ones vector. Importantly, when a new task arrives, $W_t$ can be reinitialized solely based
on the updated class count, without requiring any data. The prototypes have equal $\ell_2$ norms and identical pair-wise angles:
\begin{equation}
\small
\mathbf{w}_i^\top \mathbf{w}_j =
\begin{cases}
1, & i=j,\\[-2pt]
-\dfrac{1}{C_t-1}, & i\neq j,
\end{cases}
\qquad i,j\in\mathcal{C}_t.
\label{eq:etf_angle_t}
\end{equation}
Based on the above ETF prototypes, we construct cosine similarity matrices
aligned with the current mini-batch. Let $B$ denote the batch size and
$\mathbf{f}_a$ the feature of the $a$-th sample with label $y_a\in\{1,\dots,C_t\}$.
We define:
\begin{equation}
\small
M_F^{a,b}
=
\frac{\langle \mathbf{f}_a, \mathbf{f}_b \rangle}
{\|\mathbf{f}_a\|_2 \cdot \|\mathbf{f}_b\|_2},
\ 
M_P^{a,b}
=
\frac{\langle \mathbf{w}_{y_a}, \mathbf{w}_{y_b} \rangle}
{\|\mathbf{w}_{y_a}\|_2 \cdot \|\mathbf{w}_{y_b}\|_2},
\label{eq:similarity_matrices_final}
\end{equation}
where $\mathbf{w}_{y_a}$ is the ETF prototype of the class corresponding to
$\mathbf{f}_a$. Thus, both $M_F$ and $M_P$ are $B\times B$ matrices with the same row/column ordering, which enables direct sample-wise angular distillation.
We first normalize each row into a probability distribution with a row-wise
softmax:
\begin{equation}
\small
P_F^{a,b}=\mathrm{softmax}_b\!\left(\frac{M_F^{a,b}}{\tau}\right),\ 
P_P^{a,b}=\mathrm{softmax}_b\!\left(\frac{M_P^{a,b}}{\tau}\right),
\end{equation}
where $\tau$ is a temperature parameter. To mitigate class-imbalance (head classes
contributing more rows than tail classes), we adopt a class-balanced aggregation.
Let $n_c$ be the number of samples of class $c$ in the current mini-batch, and
$\mathcal{C}_B=\{\,c\in\mathcal{C}_{t}\mid n_c>0\,\}$ the set of present classes.
The GSA loss is computed by averaging the row-wise KL divergence \emph{per class}
and then averaging over classes:
\begin{equation}
\mathcal{L}_{\mathrm{GSA}}
=
\frac{1}{|\mathcal{C}_B|}
\sum_{c\in\mathcal{C}_B}
\frac{1}{n_c}
\sum_{a:\,y_a=c}
\mathrm{KL}\!\big(P_F^{a,:}\,\|\,P_P^{a,:}\big).
\label{eq:pad_cb}
\end{equation}
This class-balanced design strengthens the alignment for tail classes during local client optimization, ensuring they receive sufficient geometric supervision. In addition, we employ a standard classification loss $\mathcal{L}_{CLS}$ to ensure the features remain discriminative and well-separated, i.e.,
\begin{equation}
\small
    \mathcal{L}_{\mathrm{CLS}} = - \textstyle \sum_{i=1}^{C_t} y_i \log \left(e^{z_i}/{ { {\textstyle \sum_{j=1}^{C_t}}  } e^{z_j}} \right)
\label{eq6}
\end{equation}
where $y_i$ denotes the one-hot encoded ground-truth label for class $i$, and $z_i = \langle \mathbf{f}, \mathbf{w}_i \rangle$ is the similarity score between the feature $\mathbf{f}$ and the prototype $\mathbf{w}_i$ of class $i$.

\begin{algorithm}[!t]
\caption{\textsc{FEAT}}
\label{alg1}
\begin{algorithmic}[1]

\STATE \textbf{Initialize:} $R$: number of communication rounds; $K$: number of clients; $T$: number of tasks; $\mathcal{D}_{k}$: local dataset on client $k$, including both current and previous tasks; $\theta_g$: global model parameters.

\FOR {each task $t = 1$ to $T$}
    \FOR {each communication round $r = 1$ to $R$}
        \FOR {each client $k = 1$ to $K$}
            \STATE Initialize local model parameters: $\theta_k \leftarrow \theta_g$
            \STATE Select a mini-batch $\zeta$ from $\mathcal{D}_k$ and update $\theta_k$ by Eq.~\ref{eq6} if $t{=}1$, otherwise by Eq.~\ref{eq15}.
            \STATE Compute the local statistics
$\{\bar{e}_H^{(T,k)},\,\bar{e}_T^{(T,k)}\}$ via Eq.~\ref{eq9}.
        \ENDFOR
    \STATE Aggregate local models via $\theta_{g} = \frac{1}{N_C} \sum_{k} \theta_{(k)}$.
    \STATE Compute the global statistics $\{\bar{e}_H^{G},\,\bar{e}_T^{G}\}$ by Eq. \ref{eq10}.
    \ENDFOR
    \STATE Replay data in the manner of Re-Fed+ or FedCBDR.
\ENDFOR
\STATE \textbf{Inference:} Employ the EGC module to remove bias via Eqs.~\ref{eq11}, \ref{eq12} and \ref{eq13}.
\end{algorithmic}
\end{algorithm}

\subsection{Energy-based Geometric Correction (EGC)}
\begin{figure}[t]
\centering
\includegraphics[width=1.0\linewidth]{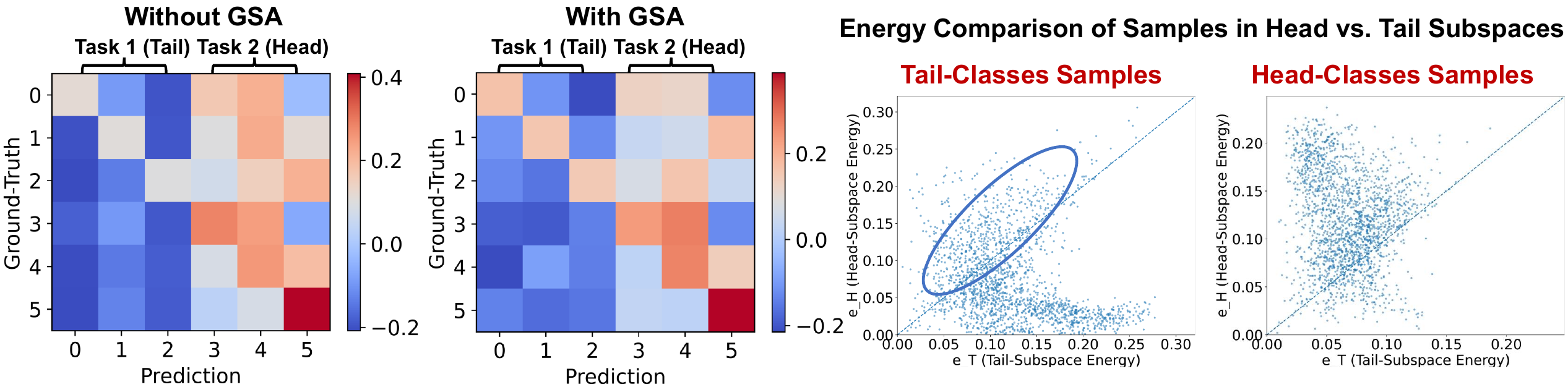}
\caption{The GSA module reduces the tendency of tail classes to drift toward head-class directions, but tail-class features still show high similarity to head-class prototypes. Moreover, a notable fraction of tail-class samples are aligned with the head-class subspace.}
\label{fig4}
\end{figure}

Although GSA mitigates cross-client feature misalignment, limited replay leaves a long-tailed distribution that induces a systematic drift of tail features toward head directions. Moreover, we measure this drift with \emph{rank-normalized subspace energies} $e_H$ and $e_T$ (Eq.~\ref{eq11}) computed via the projectors in Eq.~\ref{eq:proj_ops_ecc}. Obviously, a large fraction of tail samples continue to have $e_H>e_T$, signaling head-subspace bias (Figure~\ref{fig4}). To address this, EGC performs a lightweight inference-time correction.
\\

\noindent\textbf{Task-wise ETF subspace partitioning.}
At incremental task $t$, we maintain the ETF prototypes
$W_t=[W_T,\,W_H]\in\mathbb{R}^{d\times C_t}$ for the currently observed class set
$\mathcal{C}_t$, where $W_H=[\mathbf{w}_c]_{c\in\mathcal{C}_H}$ and
$W_T=[\mathbf{w}_c]_{c\in\mathcal{C}_T}$ correspond to head (current) and tail
(previous) classes, respectively. Following the simplex-ETF property,
\begin{equation}
\mathrm{rank}(W_H)=|\mathcal{C}_H|-1,\qquad 
\mathrm{rank}(W_T)=|\mathcal{C}_T|-1.
\end{equation}
We compute orthogonal projection operators using the Moore--Penrose pseudoinverse ($\dagger$):
\begin{equation}
\resizebox{0.9\linewidth}{!}{$
P_H = W_H (W_H^\top W_H)^{\dagger} W_H^\top,\qquad
P_T = W_T (W_T^\top W_T)^{\dagger} W_T^\top.
$}
\label{eq:proj_ops_ecc}
\end{equation}

\noindent\textbf{Training-time tail priors.}
During local client training with exemplar replay, we estimate the typical
projection energies of tail samples over the two subspaces. For a normalized
feature $\tilde{f}_a$ from replayed tail data, we maintain Exponential Moving Averages (EMA) of the rank-normalized energies:
\begin{equation}
\small
\begin{aligned}
\bar{e}_H^{(T)} &\leftarrow (1-\rho)\,\bar{e}_H^{(T)}
+ \rho \frac{1}{n_T}
\sum_{a \in \mathcal{D}_T}
\frac{\|P_H \tilde{f}_a\|_2^2}{|\mathcal{C}_H|-1},
\\
\bar{e}_T^{(T)} &\leftarrow (1-\rho)\,\bar{e}_T^{(T)}
+ \rho \frac{1}{n_T}
\sum_{a \in \mathcal{D}_T}
\frac{\|P_T \tilde{f}_a\|_2^2}{|\mathcal{C}_T|-1}.
\end{aligned}
\label{eq9}
\end{equation}
After each local training round, clients upload only the scalar EMA statistics
$\{\bar{e}_H^{(T,k)},\,\bar{e}_T^{(T,k)}\}$ to the server.
The server performs a sample-size–weighted aggregation:
\begin{equation}
\resizebox{0.88\linewidth}{!}{$
\bar{e}_H^{(G)}=\frac{1}{\sum_k n_T^{(k)}}\sum_k n_T^{(k)} \bar{e}_H^{(T,k)},
\qquad
\bar{e}_T^{(G)}=\frac{1}{\sum_k n_T^{(k)}}\sum_k n_T^{(k)} \bar{e}_T^{(T,k)},
$}
\label{eq10}
\end{equation}
where $k$ indexes clients and $n_T^{(k)}$ is the number of replayed tail
samples used for the local EMA update.

\noindent\textbf{Inference-time decontamination.}
During inference, for any normalized feature $\tilde{x}$, we compute rank-normalized
projection energies:
\begin{equation}
\small
e_H(\tilde{x})=\frac{\|P_H \tilde{x}\|_2^2}{|\mathcal{C}_H|-1},\qquad
e_T(\tilde{x})=\frac{\|P_T \tilde{x}\|_2^2}{|\mathcal{C}_T|-1}.
\label{eq11}
\end{equation}
A confidence gate is derived from the deviation above the global tail prior:
\begin{equation}
\small
g(\tilde{x})=
\max\!\left\{
\frac{e_H(\tilde{x})-\bar{e}_H^{(G)}}
     {e_H(\tilde{x})+e_T(\tilde{x})+\varepsilon},
\ 0
\right\}.
\label{eq12}
\end{equation}
The corrected representation suppresses the head-aligned component and enhances
the tail-aligned counterpart, followed by $\ell_2$-normalization:
\begin{equation}
\tilde{x}' =
\tilde{x}
- g(\tilde{x}) P_H\tilde{x}
+ g(\tilde{x}) P_T\tilde{x},
\qquad
\tilde{x}'\leftarrow \frac{\tilde{x}'}{\|\tilde{x}'\|_2}.
\label{eq13}
\end{equation}
Finally, prediction is made with ETF-based similarities:
\begin{equation}
z_c=(\tilde{x}')^\top \mathbf{w}_c,\qquad 
c \in \mathcal{C}_t.
\label{eq:final_pred_ecc}
\end{equation}

\subsection{Training Strategy}
\methodname{} adopts a stage-dependent optimization scheme:
\begin{itemize}[leftmargin=10pt]
    \item \textbf{Initial task ($t=1$):} Only the classification loss is applied:
    $\mathcal{L} = \mathcal{L}_{\text{CLS}}.$
    \item \textbf{Subsequent tasks ($t>1$):} Both the classification loss and the
    GSA loss are employed:
    \begin{equation}
    \mathcal{L}
    = \mathcal{L}_{\text{CLS}}
    + \lambda \cdot \mathcal{L}_{\text{GSA}},
    \label{eq15}
    \end{equation}
\end{itemize}
where $\lambda$ is a balancing hyperparameter.

%% file: sec/5_experiments.tex
\section{Experiments}
\subsection{Experiment Settings}
\paragraph{Datasets.} 

\begin{table*}[t]
\centering
\caption{Performance comparison between \methodname{} and baselines across three datasets under varying levels of data heterogeneity ($\beta$) and different numbers of clients ($K$). Results are averaged over three random seeds, with standard deviations reported. 
}
\setlength{\tabcolsep}{0.5mm}
\fontsize{8}{9.8}\selectfont 
\begin{tabular}{c|cc|cc|cc|cc|cc|cc}
\hline
\multirow{4}{*}{\textbf{Method}} 
& \multicolumn{4}{c|}{\textbf{CIFAR10}} 
& \multicolumn{4}{c|}{\textbf{CIFAR100}} 
& \multicolumn{4}{c}{\textbf{TinyImageNet-Subset}} 
\\\cline{2-13}
& \multicolumn{2}{c|}{3 Tasks}  & \multicolumn{2}{c|}{5 Tasks} & \multicolumn{2}{c|}{5 Tasks} & \multicolumn{2}{c|}{10 Tasks} & \multicolumn{2}{c|}{5 Tasks} & \multicolumn{2}{c}{10 Tasks} \\\cline{2-13}
& $\beta{=}0.5$ & $\beta{=}1.0$ 
& $\beta{=}0.5$ & $\beta{=}1.0$  
& $\beta{=}0.1$ & $\beta{=}0.5$ 
& $\beta{=}0.1$ & $\beta{=}0.5$  
& $\beta{=}0.1$ & $\beta{=}0.5$ 
& $\beta{=}0.1$ & $\beta{=}0.5$  
\\
\hline
& \multicolumn{12}{c}{\textbf{$K=5$ Clients}} \\\cline{1-13}
FedEWC & 40.48$_{\pm 1.6}$ & 43.11$_{\pm 3.1}$ & 22.16$_{\pm 2.1}$ & 29.38$_{\pm 2.3}$ & 18.66$_{\pm 2.6}$ & 22.64$_{\pm 4.4}$ & 10.12$_{\pm 0.6}$ & 11.34$_{\pm 0.6}$ & 11.52$_{\pm 1.4}$ & 14.27$_{\pm 2.2}$ & 7.92$_{\pm 0.7}$ & 8.85$_{\pm 0.9}$  \\
FedLwF  & 56.84$_{\pm 1.2}$ & 58.97$_{\pm 3.1}$ & 39.21$_{\pm 1.9}$ & 52.32$_{\pm 2.7}$ & 34.45$_{\pm 2.0}$ & 36.34$_{\pm 2.6}$ & 19.36$_{\pm 1.4}$ & 24.91$_{\pm 0.9}$ & 16.78$_{\pm 0.8}$ & 21.87$_{\pm 0.7}$ & 12.16$_{\pm 0.6}$ & 13.11$_{\pm 0.3}$ \\
TARGET & 43.53$_{\pm 3.8}$ & 54.87$_{\pm 4.2}$ & 36.02$_{\pm 2.4}$ & 48.91$_{\pm 1.6}$ & 30.62$_{\pm 2.9}$ & 32.89$_{\pm 3.7}$ & 13.04$_{\pm 1.1}$ & 20.57$_{\pm 0.7}$ & 17.69$_{\pm 1.5}$ & 21.11$_{\pm 1.3}$ & 11.98$_{\pm 0.5}$ & 15.28$_{\pm 0.7}$  \\
LANDER & 54.41$_{\pm 2.7}$ & 59.88$_{\pm 1.9}$ & 39.63$_{\pm 2.0}$ & 57.55$_{\pm 2.7}$ & 43.59$_{\pm 2.9}$ & 48.39$_{\pm 3.3}$ & 27.36$_{\pm 1.1}$ & 32.64$_{\pm 2.8}$ & 11.91$_{\pm 1.2}$ & 24.77$_{\pm 2.5}$ & 12.82$_{\pm 0.9}$ & 14.93$_{\pm 1.2}$  \\
Re-Fed+ & 52.95$_{\pm 4.1}$ & 61.64$_{\pm 3.9}$ & 54.15$_{\pm 2.7}$ & 58.94$_{\pm 3.2}$ & 31.92$_{\pm 4.3}$ & 37.68$_{\pm 2.3}$ & 28.64$_{\pm 0.8}$ & 38.62$_{\pm 1.7}$ & 22.89$_{\pm 0.4}$ & 26.07$_{\pm 0.5}$ & 16.71$_{\pm 1.3}$ & 21.02$_{\pm 1.4}$  \\
FedSSI & 51.44$_{\pm 3.1}$ & 56.33$_{\pm 3.1}$ & 52.26$_{\pm 2.5}$ & 56.49$_{\pm 2.5}$ & 31.27$_{\pm 1.3}$ & 37.63$_{\pm 1.3}$ & 26.71$_{\pm 1.3}$ & 31.16$_{\pm 1.3}$ & 18.56$_{\pm 1.1}$ & 21.37$_{\pm 1.1}$ & 14.41$_{\pm 0.9}$ & 17.22$_{\pm 0.9}$  \\
FedCBDR & 63.32$_{\pm 1.6}$ & 65.88$_{\pm 2.1}$ & 61.77$_{\pm 1.5}$ & 64.79$_{\pm 1.1}$ & 45.84$_{\pm 2.0}$ & 50.33$_{\pm 2.3}$ & 44.52$_{\pm 1.7}$ & 45.96$_{\pm 1.3}$ & 25.22$_{\pm 1.1}$ & 26.38$_{\pm 1.6}$ & 19.03$_{\pm 0.7}$ & 21.98$_{\pm 0.5}$  \\\hline
\methodname{}$_R$  & 60.18$_{\pm 1.4}$ & 70.28$_{\pm 1.6}$ & 60.38$_{\pm 1.5}$ & 64.62$_{\pm 1.6}$ & 37.14$_{\pm 1.8}$ & 42.31$_{\pm 1.4}$ & 32.91$_{\pm 1.3}$ & 43.34$_{\pm 1.9}$ & 23.31$_{\pm 1.5}$ & 27.36$_{\pm 1.6}$ & 19.26$_{\pm 1.7}$ & 22.74$_{\pm 1.1}$  \\
\methodname{}$_F$ & \textbf{72.67}$_{\pm 1.5}$ & \textbf{74.21}$_{\pm 2.6}$ & \textbf{70.19}$_{\pm 1.8}$ & \textbf{73.63}$_{\pm 2.0}$ & \textbf{50.14}$_{\pm 1.2}$ & \textbf{53.31}$_{\pm 0.9}$ & \textbf{48.16}$_{\pm 1.6}$ & \textbf{49.18}$_{\pm 0.8}$ & \textbf{25.88}$_{\pm 1.6}$ & \textbf{29.31}$_{\pm 0.6}$ & \textbf{19.41}$_{\pm 1.4}$ & \textbf{23.45}$_{\pm 0.4}$  \\
\hline
& \multicolumn{12}{c}{\textbf{$K=10$ Clients}} \\\cline{1-13}

FedEWC & 35.11$_{\pm 3.5}$ & 43.32$_{\pm 2.3}$ & 22.13$_{\pm 2.0}$ & 24.97$_{\pm 2.8}$ & 16.95$_{\pm 0.4}$ & 17.21$_{\pm 0.4}$ & 5.41$_{\pm 1.2}$ & 12.28$_{\pm 0.6}$ & 10.73$_{\pm 1.3}$ & 14.37$_{\pm 2.4}$ & 5.42$_{\pm 1.1}$ & 6.47$_{\pm 0.6}$  \\
FedLwF & 49.72$_{\pm 2.0}$ & 47.51$_{\pm 3.1}$ & 41.63$_{\pm 2.4}$ & 48.35$_{\pm 3.9}$ & 28.76$_{\pm 1.7}$ & 36.35$_{\pm 1.9}$ & 12.55$_{\pm 1.3}$ & 18.91$_{\pm 0.8}$ & 17.34$_{\pm 0.6}$ & 21.79$_{\pm 1.7}$ & 5.28$_{\pm 1.1}$ & 11.04$_{\pm 1.0}$ \\
TARGET & 39.20$_{\pm 2.6}$ & 40.18$_{\pm 1.2}$ & 22.45$_{\pm 1.9}$ & 27.79$_{\pm 1.5}$ & 17.89$_{\pm 1.4}$ & 22.67$_{\pm 1.7}$ & 12.72$_{\pm 1.2}$ & 15.98$_{\pm 1.1}$ & 19.12$_{\pm 1.6}$ & 25.88$_{\pm 0.5}$ & 6.23$_{\pm 1.4}$ & 10.74$_{\pm 1.0}$ \\
LANDER & 42.38$_{\pm 3.0}$ & 44.89$_{\pm 1.9}$ & 26.63$_{\pm 4.2}$ & 31.47$_{\pm 2.8}$ & 29.92$_{\pm 2.5}$ & 42.83$_{\pm 2.7}$ & 11.60$_{\pm 1.4}$ & 26.49$_{\pm 1.5}$ & 15.68$_{\pm 2.7}$ & 21.90$_{\pm 2.1}$ & 10.38$_{\pm 1.4}$ & 10.68$_{\pm 1.2}$ \\
Re-Fed+ & 44.92$_{\pm 2.2}$ & 52.97$_{\pm 3.9}$ & 37.84$_{\pm 1.8}$ & 40.16$_{\pm 1.6}$ & 32.77$_{\pm 1.9}$ & 35.84$_{\pm 0.4}$ & 28.67$_{\pm 2.2}$ & 32.74$_{\pm 0.6}$ & 20.26$_{\pm 1.7}$ & 25.67$_{\pm 1.8}$ & 15.46$_{\pm 1.3}$ & 19.32$_{\pm 1.6}$ \\
FedSSI & 41.93$_{\pm 3.1}$ & 48.27$_{\pm 3.1}$ & 38.22$_{\pm 2.5}$ & 40.45$_{\pm 2.5}$ & 30.56$_{\pm 1.3}$ & 35.72$_{\pm 1.3}$ & 27.54$_{\pm 1.3}$ & 33.16$_{\pm 1.3}$ & 16.55$_{\pm 1.1}$ & 20.27$_{\pm 1.1}$ & 11.32$_{\pm 0.9}$ & 14.49$_{\pm 0.9}$  \\
FedCBDR & 59.27$_{\pm 1.9}$ & 61.75$_{\pm 1.1}$ & 52.00$_{\pm 2.5}$ & 59.91$_{\pm 1.0}$ & 41.96$_{\pm 1.5}$ & 47.54$_{\pm 1.9}$ & 37.11$_{\pm 1.0}$ & 43.59$_{\pm 2.4}$ & 22.95$_{\pm 0.5}$ & 25.45$_{\pm 1.9}$ & 17.61$_{\pm 1.1}$ & 19.54$_{\pm 1.0}$ \\\hline
\methodname{}$_R$ & 54.62$_{\pm 1.5}$ & 62.34$_{\pm 2.1}$ & 46.59$_{\pm 1.3}$ & 49.53$_{\pm 1.1}$ & 35.75$_{\pm 1.8}$ & 40.43$_{\pm 1.6}$ & 32.38$_{\pm 1.1}$ & 37.79$_{\pm 1.4}$ & 23.11$_{\pm 1.9}$ & 27.15$_{\pm 1.1}$ & 18.16$_{\pm 1.0}$ & 20.79$_{\pm 0.9}$  \\
\methodname{}$_F$ & \textbf{70.63}$_{\pm 1.7}$ & \textbf{72.89}$_{\pm 1.9}$ & \textbf{65.48}$_{\pm 2.1}$ & \textbf{69.76}$_{\pm 1.4}$ & \textbf{45.72}$_{\pm 1.6}$ & \textbf{50.36}$_{\pm 1.4}$ & \textbf{39.36}$_{\pm 1.8}$ & \textbf{47.81}$_{\pm 2.1}$ & \textbf{25.15}$_{\pm 0.9}$ & \textbf{28.85}$_{\pm 1.0}$ & \textbf{20.49}$_{\pm 1.6}$ & \textbf{23.17}$_{\pm 1.1}$  \\
\hline
\end{tabular}
\label{tab2}
\end{table*}

Following prior work \cite{zhang2023target,huang2025soft,liu2023cross,li2024towards}, we use three widely used benchmarks: CIFAR10/100 \cite{krizhevsky2009learning}, and TinyImageNet-Subset \cite{le2015tiny}. To emulate non-iid client environments, we partition local datasets using a Dirichlet distribution. Specifically, we consider two client settings (5 and 10 clients) for all datasets, and two task granularities per dataset: CIFAR10 is split into 3 and 5 tasks with $\beta = \{0.5, 1.0\}$, while CIFAR100 and TinyImageNet-Subset are split into 5 and 10 tasks with $\beta = \{0.1, 0.5\}$.
\vspace{-0.6cm}
\paragraph{Evaluation Metric.}
Following prior work \cite{li2024towards,qiglobal,chen2023class,liao2025federated}, we report Top-1 Accuracy, defined as $N_{\text{correct}} / N_{\text{total}}$, where $N_{\text{correct}}$ and $N_{\text{total}}$ denote the numbers of correct predictions and samples, respectively.
\vspace{-0.3cm}
\paragraph{Implementation Details.}
We use ResNet-18 as the backbone for all datasets, and the classifier expands as new classes arrive. Each client trains for 2 local epochs per communication round with batch size 128. We run 100 communication rounds per task using SGD with learning rate 0.04 and weight decay $1\times10^{-5}$. The exemplar budget per task is 450/300 for CIFAR10 (3/5 tasks), 1000/500 for CIFAR100 (5/10 tasks), and 500/250 for TinyImageNet-Subset (5/10 tasks). For GSA, the temperature $\tau$ is chosen from $\{0.07, 0.5\}$ and the weight $\lambda$ from $\{0.05, 0.1, 0.5\}$. For EGC, the EMA decay $\rho$ is selected from $\{0.5, 0.7, 0.9\}$. We set the numerical stability term $\varepsilon$ to $10^{-8}$.

\subsection{Performance Comparison} 
To validate the effectiveness of the method \methodname{}, we compare it with seven state-of-the-art baselines, including FedEWC \cite{kirkpatrick2017overcoming}, FedLwF \cite{li2017learning}, TARGET \cite{zhang2023target}, LANDER \cite{tran2024text}, FedSSI \cite{li2024rehearsal}, Re-Fed+ \cite{li2025re}, and FedCBDR \cite{qi2025class}. To explore the adaptability of \methodname{} in different frameworks, we construct two hybrid versions: \methodname{}$_R$ and \methodname{}$_F$, which correspond to incorporating \methodname{} into Re-Fed+ and FedCBDR, respectively. Table \ref{tab2} summarizes the outcomes:

\begin{itemize}[leftmargin=7pt]
\item \methodname{}$_F$ consistently achieves the highest Top-1 accuracy across all cases, including various levels of heterogeneity, and different task partitions, demonstrating the robustness of the proposed de-biased framework.
\item Both \methodname{}$_R$ and \methodname{}$_F$ achieve performance improvements over their corresponding baselines, which validates the importance of the core idea behind our method, which also highlights the plug-and-play nature.
\item As the setting becomes more challenging, with stronger heterogeneity and more clients, \methodname{}’s advantage over other methods does not shrink. This indicates that its gains are not tied to easy cases but remain stable when scaled to more complex and realistic FCL scenarios.
\item Although generative replay method LANDER achieves competitive performance compared to exemplar replay-based methods in some cases, they rely heavily on a large number of generated samples, which poses potential limitations when deployed in large-scale environments.
\end{itemize}

\begin{table}[t]
\centering
\caption{Ablation studies are conducted on CIFAR-10 with 3 tasks under $\beta \in \{0.5, 1.0\}$ and on CIFAR-100 with 5 tasks under $\beta \in \{0.1, 0.5\}$. The number of clients is 5.}  
\renewcommand{\arraystretch}{1.0}
\setlength{\tabcolsep}{1.5pt} 
\fontsize{8}{10}\selectfont 
\begin{tabular}{cc|cc|cc}
\hline
& \multirow{2}{*}{\textbf{Method}} & \multicolumn{2}{c|}{\textbf{CIFAR10}} & \multicolumn{2}{c}{\textbf{CIFAR100}} \\\cline{3-6}
& & $\beta{=}0.5$ & $\beta{=}1.0$ & $\beta{=}0.1$ & $\beta{=}0.5$   \\
\hline
& FedCBDR & 63.32$_{\pm 1.6}$ & 65.88$_{\pm 2.1}$ & 45.84$_{\pm 2.0}$ & 50.33$_{\pm 2.3}$  \\
& +ETF & 62.17$_{\pm 1.6}$ & 63.66$_{\pm 2.0}$ & 44.58$_{\pm 1.9}$ & 48.69$_{\pm 1.2}$   \\
& +ETF+GSA & 68.54$_{\pm 1.6}$ & 70.42$_{\pm 1.7}$ & 47.77$_{\pm 1.2}$ & 52.23$_{\pm 1.6}$  \\
& +ETF+EGC & 69.12$_{\pm 1.8}$ & 70.83$_{\pm 2.1}$ & 47.16$_{\pm 1.4}$ & 51.72$_{\pm 1.5}$   \\
& +ETF+GSA+EGC & \textbf{72.67}$_{\pm 1.5}$ & \textbf{74.21}$_{\pm 2.6}$ & \textbf{50.14}$_{\pm 1.2}$ & \textbf{53.31}$_{\pm 0.9}$   \\
\hline
\end{tabular}
\label{tab3}
\end{table}

\subsection{Ablation Study}
This section aims to investigate the contributions of key components, including the Geometric Structure Alignment (GSA) module and the Energy-based Geometric Correction (EGC) module. The results are presented in Table~\ref{tab3}:

\begin{itemize}[leftmargin=7pt]
\item Relying solely on the ETF classifier may not yield performance gains, as severe data imbalance substantially undermines cross-source representation alignment.

\item The integration of GSA consistently enhances performance under all datasets and task settings. This improvement stems from GSA’s capacity to maintain inter-class geometric consistency, which helps reduce the representation gap among heterogeneous clients.

\item By addressing task-level imbalance during inference, the EGC module stabilizes predictions, improves performance, and normalizes decision boundaries, reducing head-task dominance. 


\end{itemize}

\begin{figure}[t]
\centering
\includegraphics[width=1.0\linewidth]{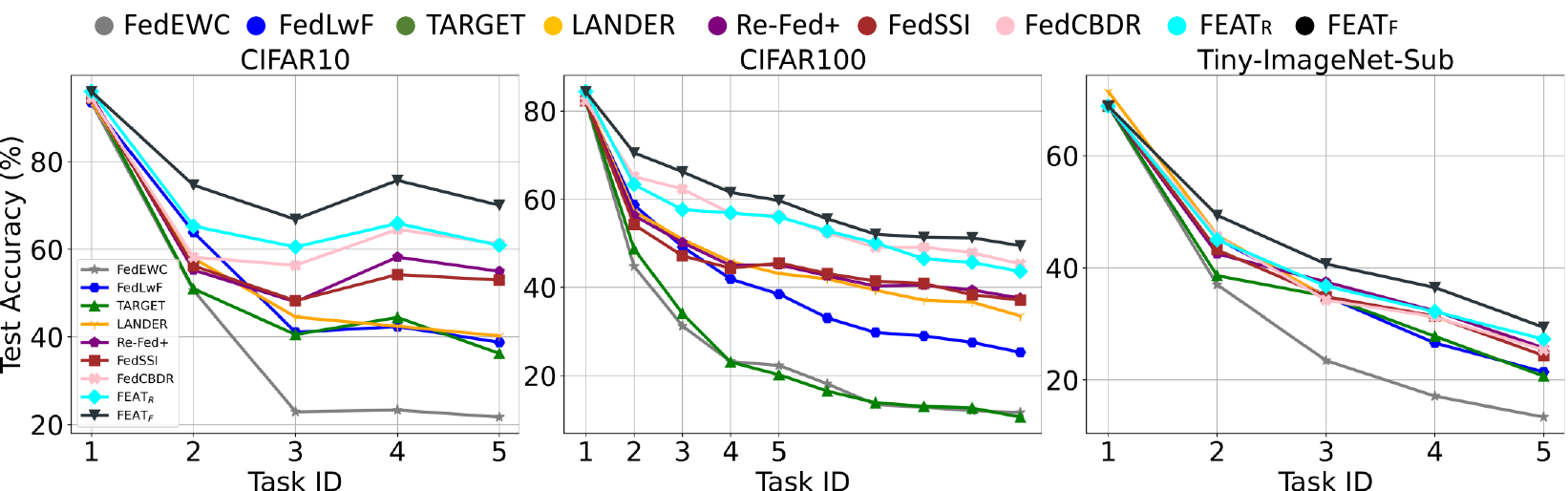}
\caption{Evaluation on CIFAR-10 (5 tasks), CIFAR-100 (10 tasks), and TinyImageNet-Subset (5 tasks) with 5 clients and heterogeneity $\beta=0.5$. \methodname{}$_F$ and \methodname{}$_R$ consistently outperform their baselines on all cases.
} 
\label{fig5}
\end{figure}
\subsection{Forgetting Curve across Incremental Tasks}
This section evaluates \methodname{} and baselines on three datasets with heterogeneity $\beta=0.5$. As shown in Figure \ref{fig5}, \methodname{}$_F$ and \methodname{}$_R$ consistently outperform their baselines across all increments by mitigating client heterogeneity and task-level imbalance. Moreover, \methodname{}$_F$ attains the highest initial accuracy and shows the slowest, most stable decline, indicating stronger long-term resistance to forgetting.

\subsection{Communication Cost Analysis}
\methodname{} retains the same training rounds and aggregation protocol as the baseline, adding only negligible overhead.
Per round, each client uploads two scalar EMA statistics ($\bar{e}_H$ and $\bar{e}_T$), which is trivial compared to model parameters, and the server performs weighted aggregation.
GSA runs locally and requires no extra transmission, while EGC is applied at inference and introduces no training-time communication.

\subsection{Analyzing the Influence of Replay Data Volume}
\begin{figure}[h]
\centering
\includegraphics[width=1.0\linewidth]{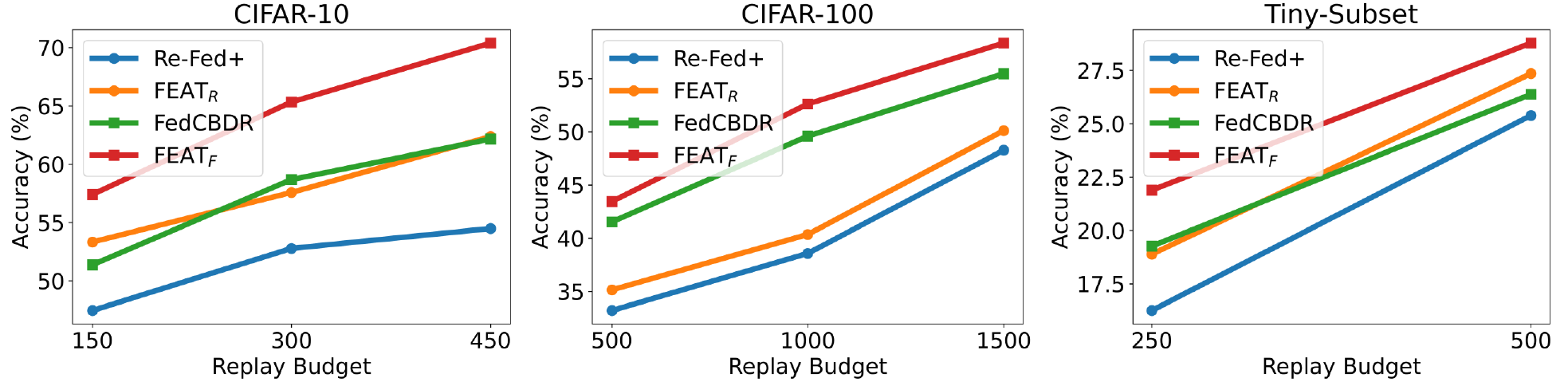}
\caption{Impact of per-task memory capacity $M$ on model performance across different datasets, where \methodname{}$_F$ shows the best overall performance.
} 
\label{fig6}
\end{figure}
\noindent This section evaluates \methodname{}$_F$ and \methodname{}$_R$ against baselines under different replay budgets on three datasets (5-task split), with replay sizes set to $\{100,150,300\}$ (CIFAR-10), $\{125,250,500\}$ (CIFAR-100), and $\{250,500\}$ (TinyImageNet-Subset).
As shown in Figure \ref{fig6}, \methodname{}$_F$ and \methodname{}$_R$ consistently achieve performance improvements over their respective baselines across all settings. These results confirm the effectiveness of our design and emphasize the importance of tackling client heterogeneity and task-level imbalance in FCL. Compared to \methodname{}$_R$, \methodname{}$_F$ shows more significant performance improvements as the replay budget increases. This can be attributed to the class-wise balanced replay inherited from FedCBDR.


\subsection{Sensitivity Study of Hyper-parameters}
\begin{figure}[h]
\centering
\includegraphics[width=1.0\linewidth]{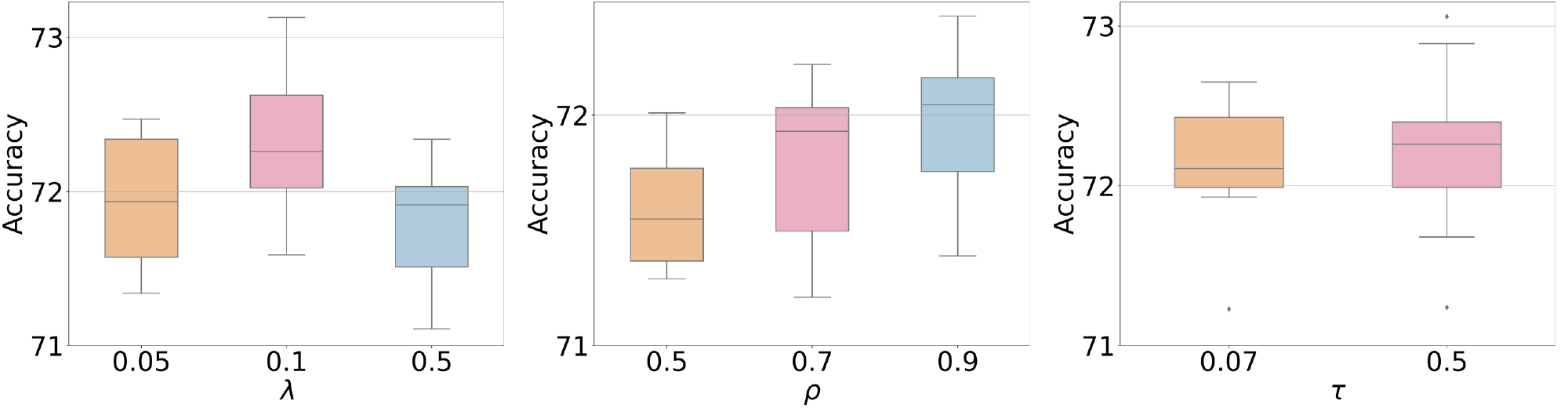}
\caption{Sensitivity analysis of hyperparameters $\lambda$, $\rho$, and $\tau$, adjusted over the ranges $\{0.05, 0.1, 0.5\}$, $\{0.5, 0.7, 0.9\}$, and $\{0.07, 0.5\}$, respectively. For each parameter, the default value is fixed while the other two are varied. It achieves robust performance under a wide range of these parameters.
} 
\label{fig7}
\end{figure}

We assess hyperparameter sensitivity on CIFAR-10 (5 tasks, $\beta{=}0.5$, 5 clients), varying $\lambda\!\in\!\{1,5,10\}$, $\rho\!\in\!\{0.5,0.7,0.9\}$, and $\tau\!\in\!\{0.1,0.5\}$. Figure~\ref{fig7} shows stable accuracy with only slight fluctuations, indicating robustness to hyperparameter choices. In general, a smaller $\lambda$ better exploits geometric structure knowledge, a larger $\rho$ yields more stable debiased learning, and $\tau$ has a minor effect. Notably, these configurations consistently outperform its baseline.


\begin{table}[h]
\centering
\caption{Performance analysis of alternative optimization strategies, including MOON, FedRCL, CLIP2FL, FedFSA, and \methodname{}, applied on top of the FedCBDR baseline. Results are reported as the mean over three runs on three datasets.}
\renewcommand{\arraystretch}{1}
\setlength{\tabcolsep}{1.5pt}
\fontsize{8}{9}\selectfont
\begin{tabular}{c|cc|cc|cc}
\hline
\multirow{2}{*}{\textbf{Methods}} & \multicolumn{2}{c|}{\textbf{CIFAR10}} & \multicolumn{2}{c|}{\textbf{CIFAR100}} & \multicolumn{2}{c}{\textbf{Tiny-Subset}} \\\cline{2-7}
 & $\beta=0.5$ & $\beta=1.0$ & $\beta=0.1$ & $\beta=0.5$ & $\beta=0.1$ & $\beta=0.5$ \\
\hline
FedCBDR & 61.77 & 64.79 & 45.84 & 50.33 & 25.22 & 26.38  \\
+ MOON & 62.56 & 66.32 & 45.14 & 49.32 & 24.73 & 26.44  \\
+ FedRCL & 58.74 & 60.21 & 42.19 & 46.46 & 20.27 & 22.46 \\
+ CLIP2FL & 62.33 & 67.97 & 46.19 & 51.11 & 24.18 & 26.67 \\
+ FedFSA & 64.42 & 68.32 & 46.88 & 51.38 & 24.61 & 26.33 \\
+ \methodname{} & \textbf{70.19} & \textbf{73.63} & \textbf{50.14} & \textbf{53.31} & \textbf{25.88} & \textbf{29.31} \\
\hline
\end{tabular}
\label{tab3}
\end{table}

\subsection{Performance Analysis of Alternative Strategies}
This section evaluates several commonly used enhancements on top of the same baseline (FedCBDR), including MOON \cite{li2021model}, FedRCL \cite{seo2024relaxed}, CLIP2FL \cite{shi2024clip}, FedFSA \cite{qi2025cross}, and \methodname{}. Table \ref{tab3} reports the mean accuracy over three runs on three datasets under a 5-task setting with 5 clients. Across all cases, \methodname{} achieves the highest accuracy when added to FedCBDR. The gains are especially clear in the more heterogeneous cases (smaller $\beta$), where other strategies provide only moderate or inconsistent improvements, while \methodname{} delivers a consistent boost. This suggests that \methodname{} is effective at handling spatiotemporal drift and cross-client imbalance, rather than only refining local features. In contrast, methods such as MOON and FedFSA offer incremental benefits but do not close the gap.

\subsection{Final Performance Comparison Across Tasks}
\begin{table}[ht]
\centering
\caption{Per-task and average accuracy (\%) of different methods on CIFAR10 and CIFAR100.}
\renewcommand{\arraystretch}{1.2}
\setlength{\tabcolsep}{1.6pt}
\resizebox{\linewidth}{!}{
\begin{tabular}{c|ccc|ccccc}
\hline
\multirow{2}{*}{} & \multicolumn{3}{c|}{CIFAR10} & \multicolumn{5}{c}{CIFAR100} \\
\cline{2-9}
& Task 1 & Task 2 & Task 3 & Task 1 & Task 2 & Task 3 & Task 4 & Task 5 \\
\hline
FedCBDR & 43.37 & 33.90 & 96.80  & 43.80 & 34.25 & 32.85 & 45.00 & 83.55 \\
+ GSA & 47.47 & 51.60 & \textbf{97.92}  & 48.75 & 36.25 & 36.10 & 52.15 & \textbf{84.40}  \\
+ EGC & 53.53 & 53.57 & 97.65  & 50.15 & 38.70 & 39.10 & 51.40 & 82.45  \\
+ EGC + GSA & \textbf{60.83} & \textbf{69.97} & 94.28 & \textbf{50.70} & \textbf{40.65} & \textbf{39.70} & \textbf{53.85} & 81.95  \\
\hline
\end{tabular}}
\label{tab4}
\end{table}
\noindent This section evaluates the final model on each task for CIFAR-10 (3 tasks) and CIFAR-100 (5 tasks) under non-IID heterogeneity $\beta{=}0.5$ with 5 clients.  As shown in Table~\ref{tab4}, adding GSA and EGC markedly improves final performance on earlier tasks, with the best results when combined. Although accuracy on the last task slightly decreases, it remains competitive, indicating stronger resistance to forgetting and more stable task-wise generalization.

\subsection{Evaluation under Different Local Epochs}
\begin{figure}[h]
\centering
\includegraphics[width=1.0\linewidth]{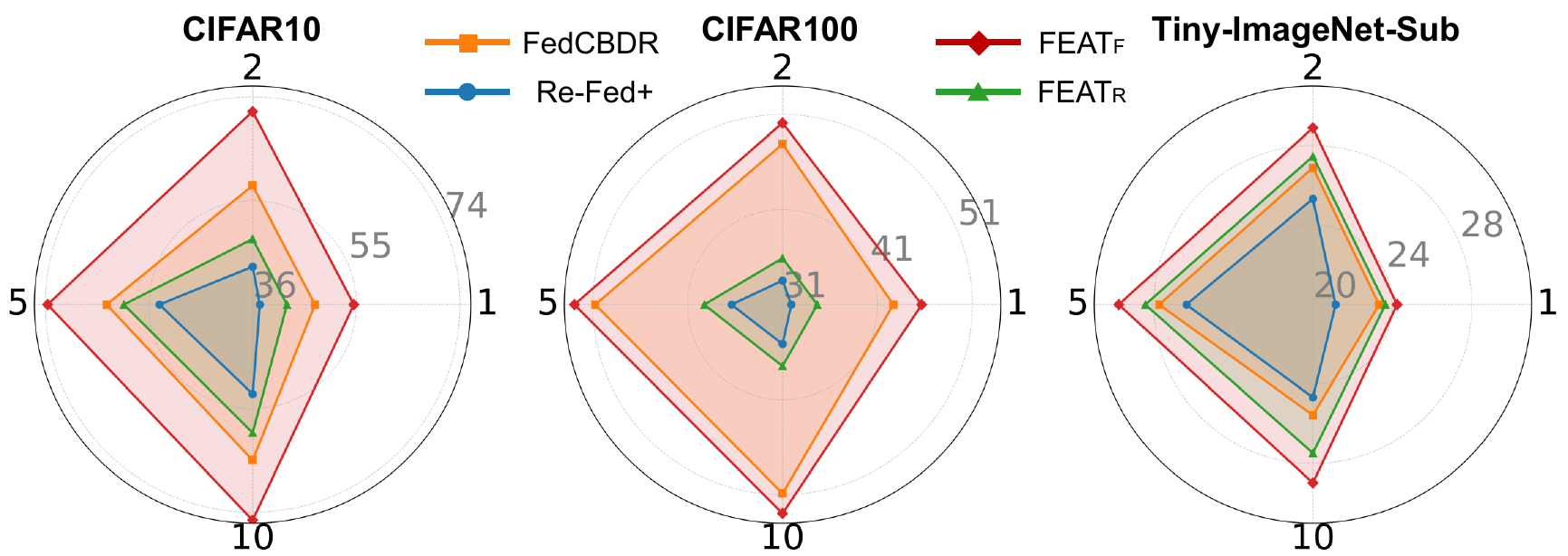}
\caption{Performance of Re-Fed+ and FedCBDR, with and without \methodname{}, across local epochs $\{1, 2, 5, 10\}$ on three datasets. And \methodname{} brings performance gains to all baselines.
} 
\label{fig8}
\end{figure}
\noindent This section evaluates the performance of Re-Fed+ and FedCBDR, with and without \methodname{}, under varying local epochs $\{1,2,5,10\}$. As shown in Figure \ref{fig8}, the versions integrated with \methodname{} consistently outperform their baselines in all settings. However, the performance improvement does not exhibit a strictly increasing trend with more local updates, indicating that the advantages of GSA and EGC may saturate or become unstable when local training dominates the optimization process. This observation provides valuable insight for future work.

\begin{figure}[h]
\centering
\includegraphics[width=1.0\linewidth]{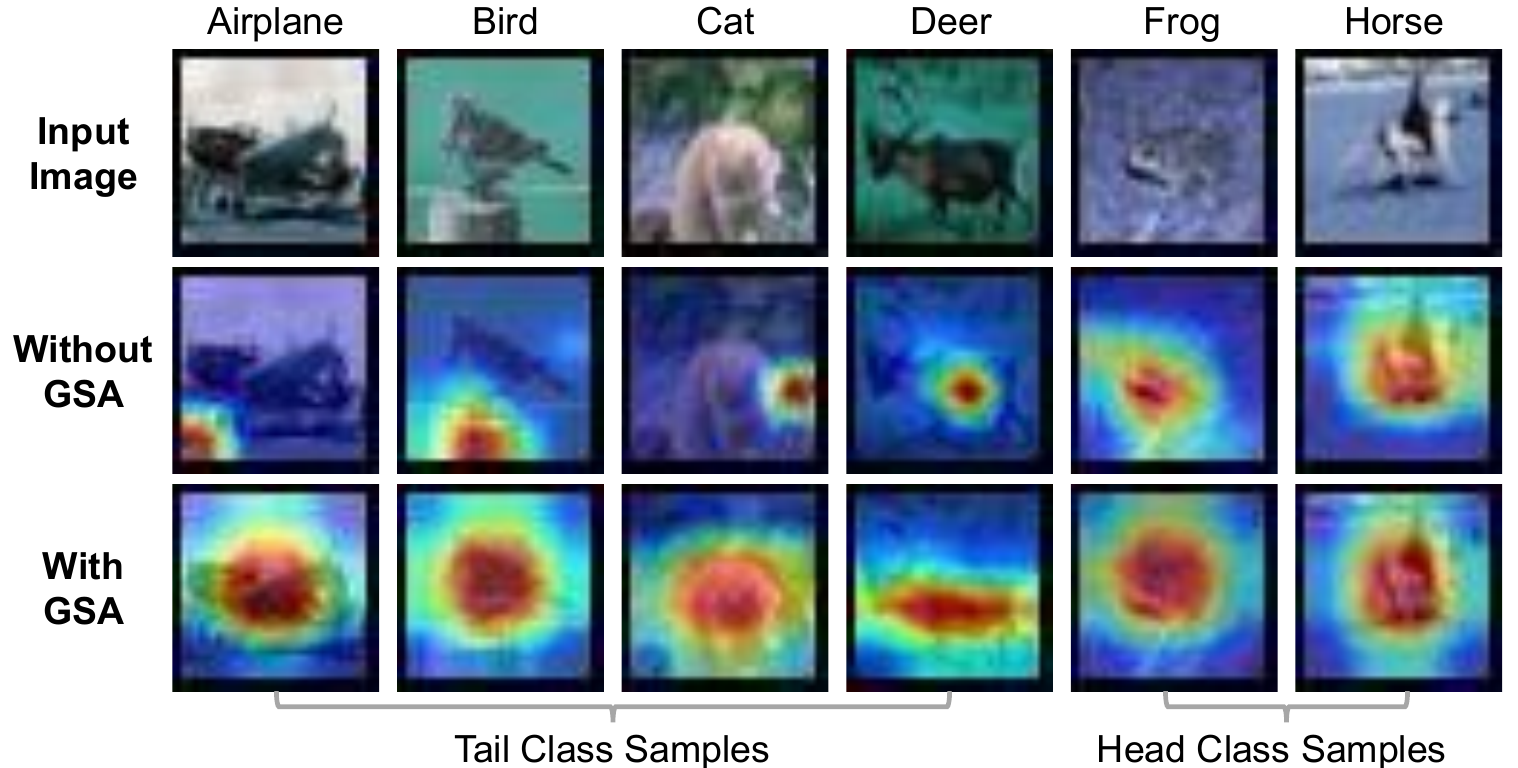}
\caption{
Visualization of visual attention. The GSA module corrects the attention bias of tail classes and strengthens the confidence of attention on head classes.
} 
\label{fig9}
\end{figure}

\subsection{Comparative Analysis of Visual Attention}
Figure~\ref{fig4} shows that GSA reduces tail-to-head drift. This section provides complementary qualitative evidence by visualizing attention maps \cite{yan2025empowering,zhang2025causality,meng2025causal,zheng2025towards,shi2026protoconnet,li2025vt} for tail and head samples. As shown in Figure \ref{fig9}, without GSA, tail classes show diffuse or off-target responses, indicating that the model is not attending reliably to the true object. With GSA, attention becomes tighter and aligned with the object region, correcting this bias. For head classes, GSA further sharpens and amplifies the focus around the object, suggesting increased confidence rather than degradation. These results demonstrate the positive impact of the GSA module.


\begin{figure}[t]
\centering
\includegraphics[width=1.0\linewidth]{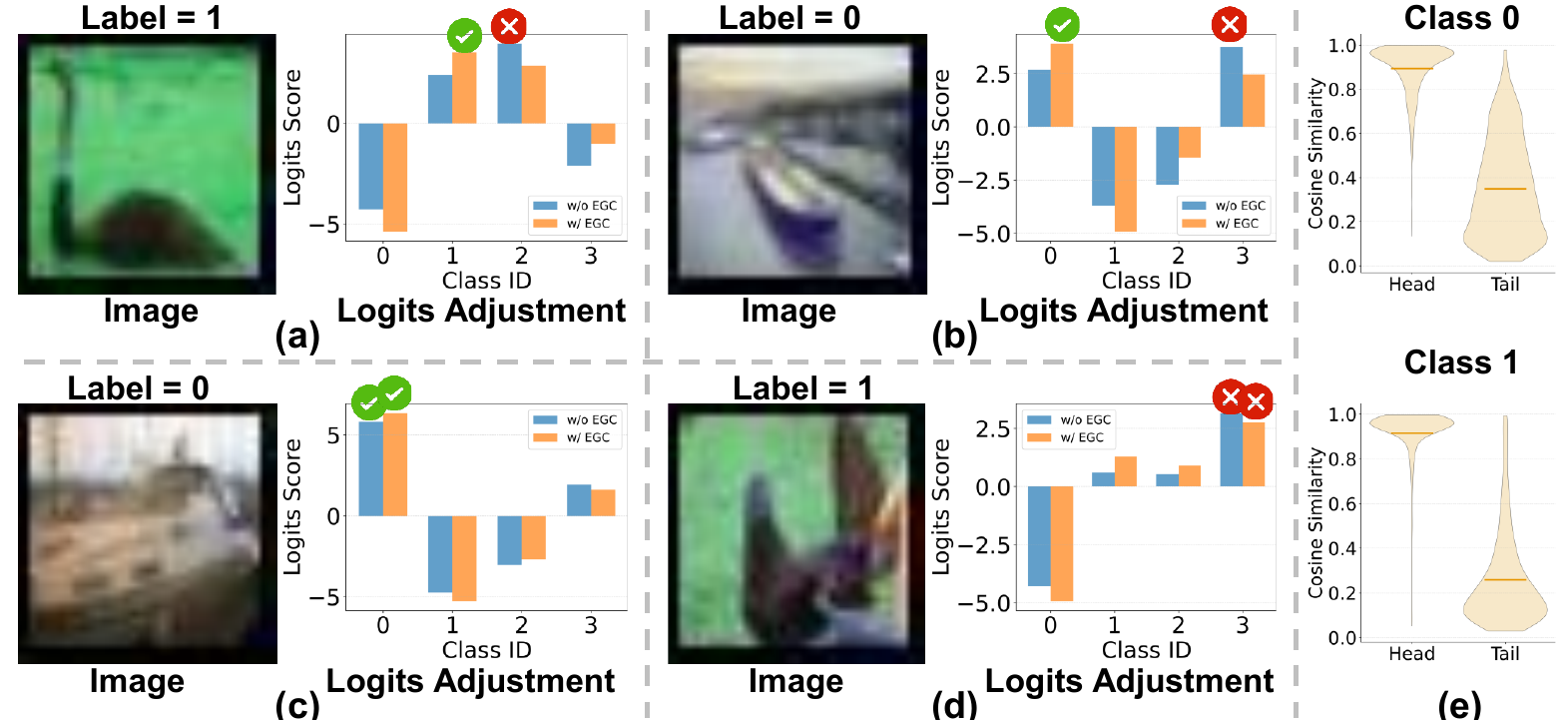}
\caption{Geometric correction effect of the EGC module on the CIFAR-10 (first two tasks). (a–b) EGC corrects the prediction bias; (c) EGC corrects the prediction bias; (d) it struggles when features deviate too far from the ground truth; (e) The removed component aligns more with the head subspace than the tail subspace for both Class 0 and Class 1.
} 
\label{fig10}
\end{figure}

\subsection{Logits-Level Behavior Analysis}
This section further analyzes how the EGC module adjusts logits at inference time. Figure \ref{fig10} shows four representative CIFAR-10 samples from the first two tasks. Obviously, cases (a) and (b) show that applying EGC amplifies the logit of the target class and suppresses the spurious head-class logits, leading to a correct prediction. Case (c) shows that EGC further enlarges the gap between the ground-truth class and the other classes. Case (d) illustrates a failure case, where the feature has drifted far from the true class manifold; in this situation, the correction is not sufficient to fully recover the correct label, although it still narrows the gap between the ground-truth class and the current top-1 class. In case (e), for tail-class samples, the removed component aligns strongly with the head-class subspace and weakly with the tail subspace, confirming that our correction targets head-attraction drift rather than a uniform shift.

%% file: sec/6_conclusion.tex
\section{Conclusions and Future Work}
To address the dual challenges of inter-client heterogeneity and task-level data imbalance, we propose \methodname{}. It consists of two key components: (1) the GSA module that enforces structural consistency between local features and globally shared prototypes, thus promoting inter-client alignment; and (2) the EGC that removes projection bias during inference, improving robustness under class-imbalanced data distributions. Experimental results show that \methodname{} consistently outperforms existing methods by effectively mitigating inter-client and inter-task bias.



\section*{Acknowledgment}

This work is supported in part by the Key Research and Development Program of Shandong Province-Innovation Capability Enhancement Program for Technology-based Small and Medium-sized Enterprises (Grant No. 2024TSGC0667); the Shandong Key Laboratory of Foundational Software (Project No. 11150004040955); the Engineering Research Center of Digital Media Technology, Ministry of Education, China; and the China Scholarship Council.